\documentclass{article}

\usepackage{arxiv}

\usepackage[utf8]{inputenc} % allow utf-8 input
\usepackage[T1]{fontenc}    % use 8-bit T1 fonts
\usepackage{hyperref}       % hyperlinks
\usepackage{url}            % simple URL typesetting
\usepackage{booktabs}       % professional-quality tables
\usepackage{amsfonts}       % blackboard math symbols
\usepackage{nicefrac}       % compact symbols for 1/2, etc.
\usepackage{microtype}      % microtypography
\usepackage{lipsum}
\usepackage{graphicx}
\graphicspath{ {./images/} }
\usepackage{hyperref}
\usepackage{comment}
\usepackage{float}
\usepackage{placeins}
\title{NEPALIGPT: A GENERATIVE LANGUAGE MODEL FOR NEPALI LANGUAGE}

\author{
 Shushanta Pudasaini\\
  Technological University Dublin\\
  Dublin, Ireland \\
  \texttt{shushanta574@gmail.com} \\
   \And
 Aman Shakya\\
  Institute of Engineering\\
  Lalitpur, Nepal \\
  \texttt{aman.shakya@ioe.edu.np} \\
  \And
 Siddhartha Shrestha \\
  Kathmandu Engineering College\\
  Kathmandu, Nepal \\
  \texttt{shrestha.siddhartha06@outlook.com} \\
  \AND
  Sahil Bhatta\\
  Fanshawe College \\
  Ontario, Canada \\
  \texttt{sahilbhatta70@gmail.com} \\
  \And
  Sunil Thapa \\
  Lambton College \\
  Ontario, Canada \\
  \texttt{sunil43thapa@gmail.com} \\
  \And
  Sushmita Palikhe\\
  Institute of Engineering \\
  Lalitpur, Nepal \\
  \texttt{081msdsa024.sushmita@pcampus.edu.np} \\  
}

\begin{document}
\maketitle
\begin{abstract}
After the release of ChatGPT, Large Language Models (LLMs) have gained huge popularity in recent days and thousands of variants of LLMs have been released. However, there isn’t any generative language model for the Nepali language, due to which other downstream tasks, including fine-tuning, have not been explored yet. To fill this research gap in the Nepali NLP space, this research proposes NepaliGPT, a generative large language model tailored specifically for the Nepali language. This research introduces an advanced corpus for the Nepali Language collected from several sources called the Devanagari Corpus. Likewise, the research introduces the first NepaliGPT benchmark dataset comprised of 4296 question-answer pairs in the Nepali language. The proposed LLM NepaliGPT achieves the following metrics in text generation: Perplexity of 26.32245, ROUGE-1 score of 0.2604, causal coherence of 81.25\%, and causal consistency of 85.41\%.

\end{abstract}

% keywords can be removed

\section{INTRODUCTION}
Large Language Models (LLMs) have proliferated, revolutionising the science of Natural Language Processing (NLP) by allowing machines to understand content produced by humans. However, such models are created for widely used languages. There are few resources for the languages used by groups of people who prefer their native language over English while using computers. One such language is Nepali.

Nepali, the official language of Nepal, is spoken by approximately 44.86\% of the country's population, according to the 2021 Nepal Population and Housing Census. Nepali is also used by communities of India, Myanmar, Bhutan and the global Nepalese diaspora\cite{nepali}. About 17.6 million people around the globe speak the Nepali language \cite{rkhanal, nlp_survey}. Nepali is based on Devanagiri script, which has 36 consonants, 13 vowels and 10 numerals \cite{vector_representation} along with several half-alphabets. There are no capital letters in the Nepali alphabet, and Nepali writing starts from left to right \cite{vector_representation}. As the same word can be written in different ways in the Nepali language, and there are multiple numbers of alphabets compared to the English language. This increases the vocabulary size which makes the Nepali language complicated for NLP tasks. 

Despite the challenges, the Nepali-specific NLP systems have been developed with a subsequent increase in accuracy. This research paper intends to create generative large language model for the Nepali language, NepaliGPT. The model can be used to get responses in the Nepali language and develop generative algorithm based applications such as chatbots, articles, question answers and so on by finetuning the language model on custom data.

\begin{figure}
    \centering
    \includegraphics[width=1\linewidth]{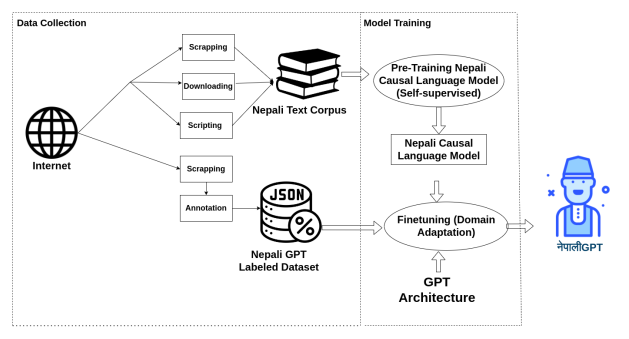}
    \caption{System Architecture for NepaliGPT: Data Collection, Model Training, and Fine-Tuning Pipeline}
    \label{fig:system_architecture}
\end{figure}

\section{ LITERATURE REVIEW}
\label{sec:headings}
In 2017, Google released a paper \textbf{Transformer is all you need} \cite{attentionisallyouneed} which changed the landscape of NLP with the introduction of the attention mechanism. Following the transformer architecture, many large pre-trained language models came to rise, such as \href{https://arxiv.org/pdf/1810.04805}{BERT}, \href{https://arxiv.org/pdf/2305.10435}{GPT}, \href{https://arxiv.org/pdf/1907.11692}{Roberta}, \href{https://arxiv.org/pdf/2302.13971}{Llama}, and \href{https://arxiv.org/pdf/2310.06825}{Mistral}.  With the advancement of such models, the development of NLP in English rises to the top, but the development of NLP in low-resource languages such as Nepali is still in the early phase. There hasn’t been any groundbreaking development in the Nepali language. There are few language models that has been developed for the Nepali Corpus major, such as \textbf{NepaliBERT} \cite{nepalibert}, and \textbf{NepBERTa} \cite{nepberta}. Here, both models were trained on BERT architecture, the first one was trained for Masked Language Modelling (MLM) along with next-word prediction, and the latter one adapting the RoBERTa’s strategy was only trained for MLM. So, they are more of Natural Language Understanding (NLU) models i.e. cannot generate Nepali tokens to communicate.

\section{Methodology}
\label{sec:others}
To build the NepaliGPT, a RAW Nepali corpus was collected by scraping from multiple available public Nepali sites and translating existing English sources into Nepali. It was further filtered by checking if any alphanumeric character existed using pre-processing steps, along with manual verification. Finally, Causal Language Model (CLM), i.e. GPT2 architecture, was trained from scratch on the collected Devanagiri Corpus.
Then, to answer user's queries, CLM on the collected Nepali Question-Answer corpus was further finetuned, which was collected by scraping. The entire process can be visualised in Figure  \ref{fig:system_architecture}.

\subsection{BUILDING NEPALI CORPUS}
\subsubsection{Data Collection}
Initially, we researched a publicly available corpus on the Nepali Language. Below are some of the major corpora we found.
\begin{itemize}
    \item Nepali National Corpus: It is a 14 million words corpus that is divided into three types of corpora i.e. written Corpus, parallel Corpus, and spoken Corpus \cite{annotation}
    \item A Large-Scale Nepali Text Corpus: It is collected by scraping from the internet \cite{nepalitext_corpus}, and has a collection of over 90 million words in Nepali. The corpus contains news articles, books, and websites. It can be downloaded freely from the IEEE DataPort website.
    \item OSCAR Corpus: OSCAR project stands for Open Super-large Crawled Aggregated Corpus. It is a multilingual, unannotated raw data corpus commonly used in the pretraining of large deep learning models. It is obtained by filtering Common Crawl Croups. OSCAR Nepali Corpus is a collection of over 100 million words of Nepali text \cite{monolingual_approach}.
\end{itemize}
Finally, we obtain our final text corpora by combining the following strategies:
\begin{itemize}
\item Scraping Nepali News Portal: Using Scrapy, a web crawling framework, and BeautifulSoup, a web scraping framework, five of the most popular online Nepali news portal i.e. ekantipur.com, nepalinews.com, onlinekhabar.com, setopati.com, and nagariknews.com were crawled, and finally, text from HTML code was extracted.
\item Downloading available raw Nepali textual data: All scraped content was combined into a single .txt file and further processed for cleaning. Finally, it was split into multiple .txt files, each containing 10,000 lines, to facilitate easier data management.
\item Scripting on open-source NLP datasets: Publicly available NLP datasets for tasks like sentiment analysis, fake news detections, etc, in both Nepali and English languages were listed. English Datasets were translated into Nepali using Google Translate and further verified and filtered using processing steps like looking for alphanumeric characters. Finally, once all Nepali datasets were prepared, each row was read and compiled into a raw, unannotated .txt file for training.
\end{itemize}

\subsubsection{CORPUS ANALYSIS}
\begin{figure}[!htbp]
    \centering
    \includegraphics[width=0.8\linewidth]{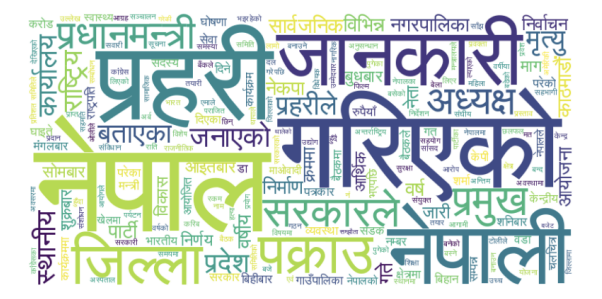}
    \caption{Wordcloud representation of the Devanagari Corpus}
    \label{fig:word_cloud_nepali}
\end{figure}

\begin{figure}[!htbp]
    \centering
    \includegraphics[width=0.8\linewidth]{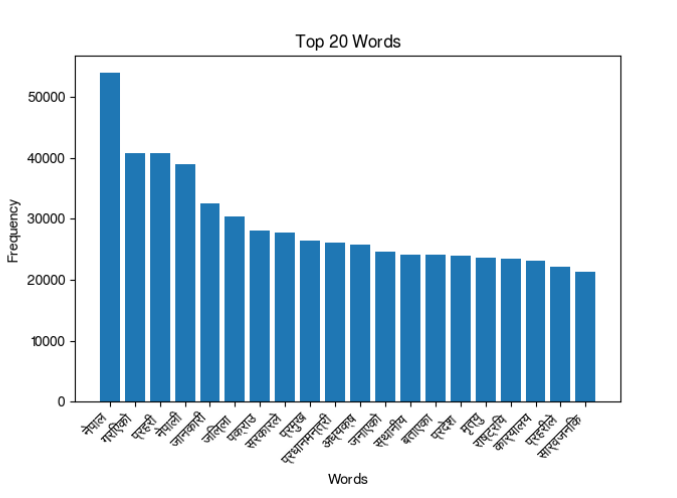}
    \caption{Bar graph of the most frequent words in the Devanagari Corpus}
    \label{fig:frequent_words_nepali}
\end{figure}

 The Nepali corpus, which we named the Devanagari Corpus, was further preprocessed to remove noise and non-Nepali tokens. To better understand the corpus and identify any potential bias, the corpus was tokenised, and the frequency of each word was calculated. The wordcloud visualising the most common words in the corpus is represented in Fig. \ref{fig:word_cloud_nepali}, and the most frequent words and their frequency are represented in Fig. \ref{fig:frequent_words_nepali}.

\FloatBarrier

We further calculated different statistics for our Nepali Corpus, like the total number of tokens, the total number of sentences, unique words, corpus size, etc. A complete table about statistics can be found in the Table \ref{data_table}:

 \begin{table}[h!]
\centering
\begin{tabular}{@{}ll@{}}
\toprule
\textbf{Nepali Corpus Statistics}                                     &                                \\ \midrule
Total number of tokens in the corpus                                  & 383303233                      \\
Total number of sentences across all files                            & 23211003                       \\
Combined average sentence length (in words)                           & 16.513859095                   \\
Combined average word length (in characters)                          & 5.7533011546                   \\
Combined vocabulary size (unique words)                               & 4494764                        \\
Size of the whole corpus                                              & 9.3 GB                         \\ \bottomrule
\end{tabular}
\centering
\caption{Basic text analysis statistics of Nepali Corpus}
\label{data_table}
\end{table}

\subsection{Building Nepali CLM}
\subsubsection{Training Byte Pair Encoding (BPE) based custom Nepali Tokeniser from scratch}
We trained a custom BPE tokeniser utilising the collected Nepali Corpus from scratch using Sentencepiece Library \cite{tokenizer_detokenizer}. The BPE algorithm \cite{data_compression_algorithm} works by initialising the vocabulary with all the bytes or characters on the corpus. Then, the frequency for each character was calculated and repeatedly merging the most common consecutive characters into the vocabulary till we reach the desired number of tokens. For the Nepali BPE tokeniser, we followed the BPE algorithm and collected the 10,000 most common consecutive characters or subwords assigned a unique ID for each subword.

A basic overview of training the custom Nepali BPE Tokenizer is shown in Figure \ref{fig:frequent_words_nepali}.
\begin{figure}[h]
\centering
    \includegraphics[width=0.8\linewidth]{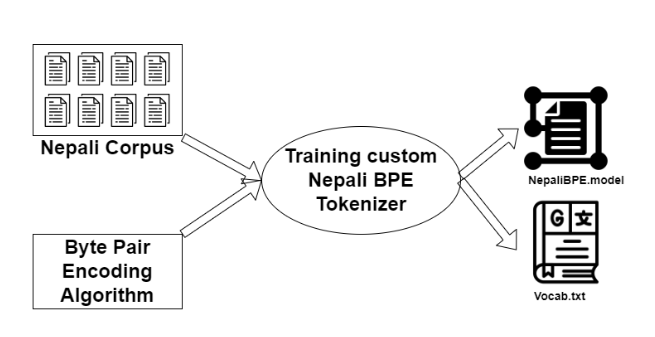}
     \caption{Block Diagram representing methodology for training the custom Nepali BPE Tokeniser}
    \label{fig:cutsom_bpe_tokenizer}
\end{figure}

\subsubsection{Pretraining CLM for Nepali Language}
Firstly, we pretrained the CLM models with our Nepali Corpus to make sure the model has a better understanding of the Nepali language. For CLM, we used the GPT2 architecture with zero weight initialisation. Using the transformers API and our custom Nepali BPE Tokenizer we pre-trained the GPT2 architecture using our collected Nepali Corpus of 13 million Nepali text corpus and learned to predict the next token provided the previous token in the sequence.

\subsection{Building Nepali Question Answer Dataset}
To make sure our Causal Language Model (CLM) model is able to answer the query provided by the user we collected the question-answer pair and listed it in the JavaScript Object Notation (JSON) format. The steps followed to build the Nepali Question Answer dataset are explained in the sections below.
\subsubsection{Scraping from Nepali Websites}
We collected question-answer pairs by scraping from different public Nepali Websites that have question answers on them, utilising the Scrapy and BeautifulSoup frameworks. We listed the question-answer pair in JSON. Below is the list of the websites we used.
\begin{itemize}
    \item \href{https://www.imnepal.com/nepali-general-knowledge-quiz-questions-answers/}{75 Nepali General Knowledge Quiz Questions Answers}
    \item \href{https://www.apsarapandey.com.np/2022/07/nepali-quiz.html}{Nepali GK and Quiz Question Answer} 
    \item \href{https://www.dhurbaghale.com.np/2023/05/150-nepali-gk-questions-and-answers.html}{150+ Nepali GK Questions and Answers | Nepali Samanya Gyan}  
\end{itemize}
\subsubsection{Translating publicly available English Question-Answering Dataset}
We collected 4000 question-answer pairs from Kaggle Datasets. Then, translated using the Google Translate API and verified by applying post-processing steps like checking if any alphanumeric character exists. Further verified manually.

\begin{figure}[!htbp]
    \centering
    \includegraphics[width=0.8\linewidth]{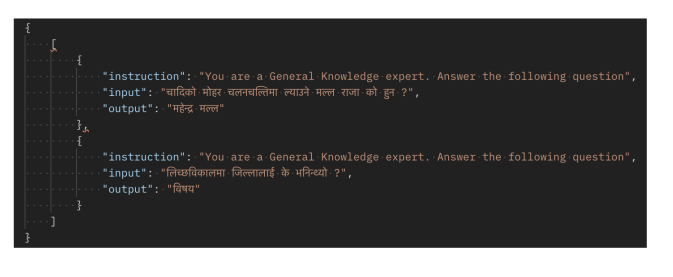}
    \caption{Data format used for Question Answering Task}
\end{figure}
\FloatBarrier
\subsection{Finetuning Nepali CLM on question answering dataset}
After we had the tokeniser, models, and supervised dataset from 3.2 and 3.3, we further finetuned the CLM model to answer questions. We updated the entire weights during this fine-tuning step, making sure the model learns to answer the user's query.

\begin{figure}[H]
    \centering
    \includegraphics[width=0.6\linewidth]{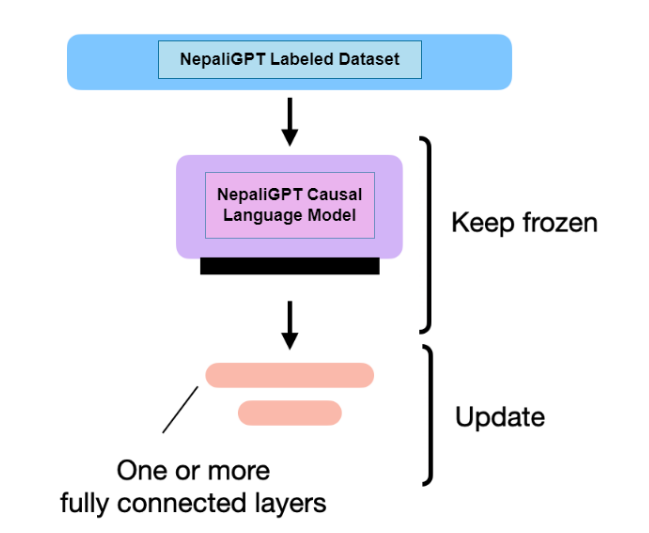}
    \caption{Finetuning Nepali Causal Language Model}
    \label{fig:enter-labe}
\end{figure}

\section{Experimental Setup}
\subsection{Hyperparameters used for Pretraining Nepali CLM}
We ran multiple experiments for building Nepali CLM i.e. one with a smaller data corpus and the other with a bigger data corpus.  For hyperparameters on both use cases, i.e. CLM and finetuning for question answering, we researched different available solutions in the English-recommended hyperparameter and set it to the optimal values.

% Table 2: Data size and training Hyperparameters
\begin{table}[h!]
\begin{tabular}{ll}
\toprule
\textbf{Hyperparameter} & \textbf{Values} \\ \midrule
Data Corpus size & 3.2 GB \\
learning rate & 5e-5 \\
train\_batch\_size & 12 \\
eval\_batch\_size & 5 \\
seed & 42 \\
optimizer & Adam \\
train\_batch\_size & 12 \\
number of epochs & 5 \\
mixed precision training & Native AMP \\
model checkpoint & GPT-2 \\ \bottomrule
\end{tabular}
\centering
\caption{Data size and training Hyperparameters Experiment I}
\end{table}

\begin{table}[!htbp]
\label{tab:hyperparams_exp2}
\begin{tabular}{ll}
\hline
\textbf{Hyperparameter}             & \textbf{Values} \\ \midrule
Data Corpus size & 9.6 GB \\
learning rate                       & 5e-5            \\
train\_batch\_size                  & 12              \\
eval\_batch\_size                   & 5               \\
seed                                & 42              \\
optimizer                           & Adam            \\
train\_batch\_size                  & 12              \\
number of epochs                    & 5               \\
mixed precision training            & Native AMP      \\
model checkpoint                    & GPT-2           \\
\hline
\end{tabular}
\centering
\caption{Data size and training hyperparameters for Experiment II}
\end{table}
\FloatBarrier

\subsection{Hyperparameter for fine-tuning Nepali CLM and building Nepali GPT}

\begin{table}[!htbp]
\begin{tabular}{ll}
\toprule
\textbf{Hyperparameter} & \textbf{Values} \\ \midrule
Base model & NepaliGPT \\
Max Length & 150 \\
Learning Rate & 0.01 \\
Weight Decay & 0.01 \\
fp16 & True \\
warmup ratio & 0.01 \\
seed & 42 \\
train batch size & 8 \\
test batch size & 8 \\
Number of epoch & 10 \\ \bottomrule
\end{tabular}
\centering
\caption{Hyperparameter Values for Finetuning NepaliGPT model}
\end{table}
\FloatBarrier

\subsection{Evaluation Metrics:}
\subsubsection{ROGUE (Recall Oriented Understanding for Gisting Evaluation) Score}
\begin{figure}[!htbp]
    \centering
    \includegraphics[width=0.6\linewidth]{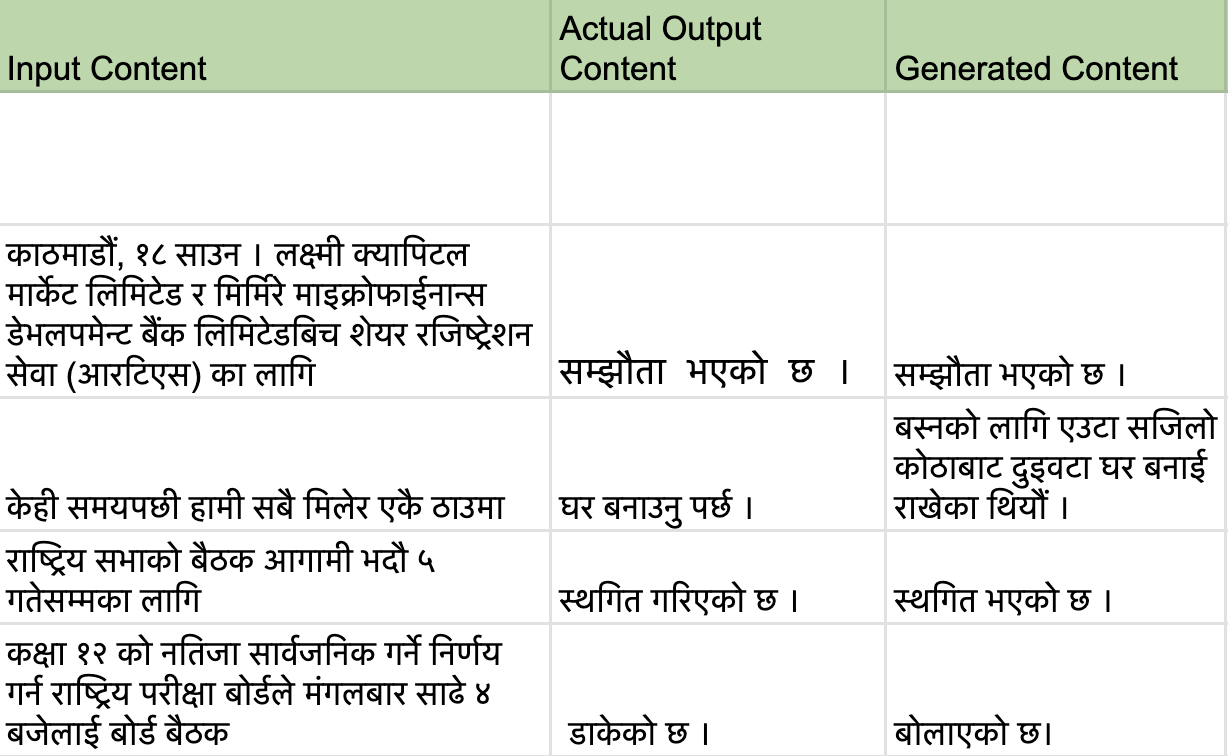}
    \caption{Data Samples Used for Evaluation}
    \label{data_samples}
\end{figure}

ROGUE is used to calculate the degree of overlap between produced outputs and reference outputs. For CLM, we removed the second half of 50 sentences and treated the removed part as a reference text, and the tokens predicted by Nepali CLM as the produced text and calculated the Rogue Score to figure out the evaluation metrics of our CLM model. Further for Nepali GPT, we have split the collected data into train, validation, and test datasets, among which the test datasets were not provided during the finetuning step. These test datasets were used to calculate the ROGUE score to evaluate the performance of Nepali GPT.
A few of the samples from the evaluation set are represented in the Fig\ref{data_samples}.

\FloatBarrier
\subsubsection{Casual Coherence and Casual Consistency}
 To measure these metrics, we used three different human evaluators, each with a different profile. The profile of the evaluators is as mentioned in the Table \ref{dataprofilesannotators}. 
\begin{table}[!htbp]
\begin{tabular}{lllll}
\toprule
\textbf{Annotator}  & \textbf{Gender} & \textbf{Education} & \textbf{Age} & \textbf{Nepali as Mother Tounge} \\ \midrule
Annotator 1 & Male   & Diploma   & 20-25  & Yes  \\
Annotator 2 & Female & Bachelors & 25-30  & No   \\
Annotator 3 & Male   & Masters   & 40-50  & Yes  \\ \bottomrule
\end{tabular}
\centering
\caption{Data Profile of the Annotators}
\label{dataprofilesannotators}
\end{table}
\FloatBarrier
 Casual Coherence means evaluating the model’s output based on whether the model is generating output that makes sense and is correct. For e.g. CLM produces “The glass fell from the table and broke”. This is a logically correct sentence, as when glass falls it breaks. So, this can be considered a causally coherent sentence. Casual Consistency refers to the consistency of that casual coherence in texts generated by CLM with one another. For e.g. we want the model not to predict that the glass is broken if the glass is grabbed before hitting the ground.
\subsubsection{Perplexity}
During the training process, perplexity was used as a major evaluation metric. Perplexity refers to how confident the model is while predicting the tokens based on the context or prior tokens. High perplexity means the model is less confident in generating the next tokens.

\section{RESULTS}
\subsection{Nepali CLM}
Experiments were performed that involved varieties of
corpus: small and large, and varieties of hyperparameters:
default, modified and fine-tuned. Among them, othe utput of the
Two of the experiments are briefly explained.

For the first experiment, i.e. with a smaller data corpus of 3.2 GB following were the obtained training metrics:

\begin{table}[!htbp]
\begin{tabular}{llll}
\toprule
\textbf{Epoch}  & \textbf{Training Loss} & \textbf{Validation Loss} & \textbf{Perplexity} \\ \midrule
1.0 & 6.23568   & 5.235684   & 105.5983  \\
2.0 & 6.22594 & 5.213568 & 104.2356   \\
3.0 & 6.21456   & 5.211457   & 97.56823  \\ 
4.0 & 6.21365   & 5.201234   & 92.56891  \\ 
5.0 & 6.20156   & 5.182356   & 87.23567  \\ \bottomrule
\end{tabular}
\centering
\caption{Training results from experiment I}
\end{table}

\begin{figure} [hbtp!]
\centering
\includegraphics[width=0.8\linewidth]{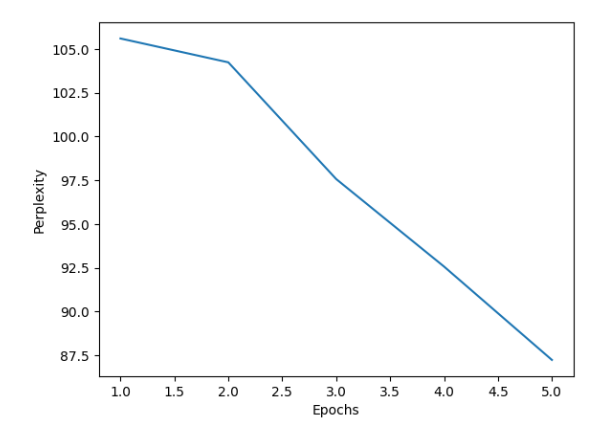}
\caption{Epoch Vs Perplexity from experiment I}
\label{fig:epoch_vs_perplexity}
\end{figure}
\FloatBarrier

 For the second experiment, i.e. with a bigger data corpus of 9.6 GB following were the obtained training metrics:
 
 \begin{table}[!htbp]
\begin{tabular}{llll}
\toprule
\textbf{Epoch}  & \textbf{Training Loss} & \textbf{Validation Loss} & \textbf{Perplexity} \\ \midrule
1.0 & 3.7654   & 3.6291   & 37.6789  \\
2.0 & 3.5857 & 3.4442 & 31.31812   \\
3.0 & 3.5051   & 3.3439   & 29.2214 \\ 
4.0 & 3.4688   & 3.3439   & 28.3294  \\ 
5.0 & 3.3968   & 3.2705   & 26.32245  \\ \bottomrule
\end{tabular}
\centering
\caption{Training results from experiment II}
\end{table}

    \begin{figure} [hbt!]
    \centering
    \includegraphics[width=0.8\linewidth]{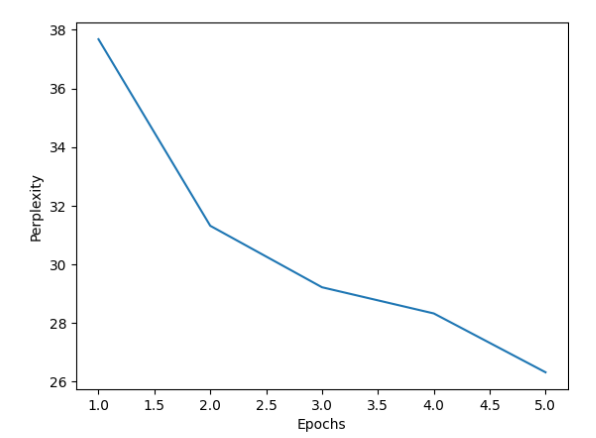}
    \caption{Epoch Vs Perplexity from experiment II}
    \label{fig:epoch_vs_perplexity_exp_2}
\end{figure}
\FloatBarrier

\subsection{NepaliGPT}

We selected 50 unseen data points from our test dataset and calculated the Rogue-N score and causal coherence, and consistency. The complete metrics are shown below. 

\begin{table}[!htbp]
\begin{tabular}{llll}
\toprule
\textbf{ROUGE Type}  & \textbf{Score} & \textbf{Validation Loss} & \textbf{Benchmark for English Dataset} \\ \midrule
ROGUE-1 & 0.2604   & 0.59 \\
ROGUE-2 & 0.1466 & 0.25   \\
ROGUE-1 & 0.1501   & 0.32   \\ \bottomrule
\end{tabular}
\centering
\caption{ROUGE Scores obtained from evaluation data for NepaliGPT}
\end{table}

\begin{table}[!htbp]
\begin{tabular}{lc}
\toprule
\textbf{Evaluation criteria}  & \textbf{Value} \\ \midrule
Causal Coherence  & 81.25 \% \\
Causal Consistency  & 85.41 \% \\ \bottomrule
\end{tabular}
\centering
\caption{Causal Consistency and Coherence from NepaliGPT model}
\end{table}
\FloatBarrier

\subsection{Training Time and Inference Time}
The pretraining of the causal language model for the development of Nepali CLM took approximately 3 days, 7 hours and 48 minutes. The model was trained on the NVIDIA Tesla T4 with 16 GB memory, 320+ GB/second bandwidth, and 2560 Nvidia CUDA cores. The Question Answering model finetuning took approximately 6 hours and 54 minutes on the same GPU device. The inference time of the NepaliGPT model is 0.7 seconds, i.e. it takes only 0.7 seconds to generate new texts given some text in any context. This inference time is recorded from the HuggingFace API endpoints.

\section{CONCLUSION AND FUTURE WORK}
 In conclusion, we were able to create a decent model by acquiring 26.32245 perplexity, which is comparable to the perplexity of the GPT-2 model on the English Wikipedia dataset, i.e. 24.35. In addition, the model received a ROUGE-1 score of 0.1501, which is equivalent to the GPT-2 model’s ROUGE-1 score of 0.148 on the English Wikipedia dataset \cite{unsupervised_multitask_learner}.  Likewise, the model can produce text that is accurate, consistent, and causally coherent based on the human review. The model’s correctness in terms of causal coherence was 81.25\%, and causal consistency was 85.41\%. The Nepali Corpus and NepaliGPT datasets developed during this research can be used to improve the results of generative-based language models. There is significant potential for further improvements. Usages of bigger and more diverse datasets, including text from different sectors so that the model has more insights on different terminologies and can be finetuned to use on different use cases. Further, evaluation of different datasets and approaches like translation+English models might provide more insights into the model’s capability. Moreover, finetuning on instruction-based datasets might lead to an instruction-following model.

\bibliographystyle{unsrt}  
\bibliography{references}  %%% Remove comment to use the external .bib file (using bibtex).


\end{document}